\documentclass[10pt,twocolumn,letterpaper]{article}

\usepackage{cvpr}              % To produce the CAMERA-READY version
\usepackage[dvipsnames]{xcolor}

\def\customparskip{.5em}
\renewcommand{\paragraph}[1]{\vspace{\customparskip}\noindent\textbf{#1}}

\definecolor{cvprblue}{rgb}{0.21,0.49,0.74}
\usepackage[pagebackref,breaklinks,colorlinks,citecolor=cvprblue]{hyperref}

\usepackage[utf8]{inputenc} % allow utf-8 input
\usepackage[T1]{fontenc}    % use 8-bit T1 fonts
\usepackage{booktabs}       % professional-quality tables
\usepackage{amsfonts}       % blackboard math symbols
\usepackage{nicefrac}       % compact symbols for 1/2, etc.
\usepackage{microtype}      % microtypography
\usepackage{xcolor}         % colors
\usepackage{lipsum}
\usepackage{bm}
\usepackage{comment}
\usepackage{algorithm}
\usepackage{algpseudocode}
\algtext*{EndWhile}% Remove "end while" text
\algtext*{EndIf}% Remove "end if" text
\algtext*{EndFor}% Remove "end if" text

\usepackage{bbding}
\usepackage{tabularx}
\usepackage{booktabs}
\usepackage{colortbl}
\usepackage{multirow}
\usepackage{xspace}
\usepackage[acronym]{glossaries}
\usepackage{float}
\usepackage{wrapfig}
\usepackage{gensymb}

\usepackage{pifont}
\def\paperID{7686} % *** Enter the Paper ID here
\def\confName{CVPR}
\def\confYear{2024}

\definecolor{tabgreen}{HTML}{E6FFE3}
\definecolor{tabred}{HTML}{FFEEEE}
\definecolor{tabyellow}{HTML}{FFFEE5}

\newcommand{\raysamp}{\boldsymbol{\mu}}
\newcommand{\colorval}{\mathbf{c}}
\newcommand{\densityval}{\tau}
\newcommand{\weightval}{w}
\newcommand{\alphaval}{\alpha}
\newcommand{\pixelcolor}{\mathbf{C}}
\newcommand{\timeval}{t}
\newcommand{\timediff}{t_d}
\newcommand{\noise}[1]{\mathbf{z}_{#1}}
\newcommand{\predimg}{\mathbf{\hat{x}}_\phi}
\newcommand{\gtimg}{\mathbf{x}}
\newcommand{\gtvideo}{\mathbf{X}}
\newcommand{\prednoise}{\boldsymbol{\epsilon}_\phi}
\newcommand{\prednoisefinetune}{\boldsymbol{\epsilon}'_\phi}
\newcommand{\gtnoise}{\boldsymbol{\epsilon}}
\newcommand{\textcond}{\mathbf{y}}
\newcommand{\params}{\theta}
\newcommand{\network}{\mathcal{N}_\params}
\newcommand{\extrinsics}{\mathbf{T}}
\newcommand{\weighting}{w(\timediff)}
\newcommand{\feature}{\mathbf{f}}
\newcommand{\msds}{\nabla_{\params}\mathcal{L}_{\text{3D}}}
\newcommand{\vsd}{\nabla_{\params}\mathcal{L}_{\text{IMG}}}
\newcommand{\vsds}{\nabla_{\params}\mathcal{L}_{\text{VID}}}
\newcommand{\parammsds}{P_\text{3D}}
\newcommand{\paramvsd}{P_\text{IMG}}

\newcommand{\textsds}{\text{3D}}
\newcommand{\textvsd}{\text{IMG}}
\newcommand{\textvsds}{\text{VID}}

\newcommand{\itersa}{N_\text{stage-1}}
\newcommand{\itersb}{N_\text{stage-2}}
\newcommand{\itersc}{N_\text{stage-3}}

\title{4D-fy: Text-to-4D Generation Using Hybrid Score Distillation Sampling}

\author{
  Sherwin Bahmani$^{1,2}$
  \space\space
  Ivan Skorokhodov$^{3,4}$
  \space\space
  Victor Rong$^{1,2}$
  \space\space
  Gordon Wetzstein$^{5}$
  \space\space
  Leonidas Guibas$^{5}$\\
  \space\space
  Peter Wonka$^{3}$
  \space\space
  Sergey Tulyakov$^{4}$
  \space\space
   Jeong Joon Park$^{6}$
  \space\space
  Andrea Tagliasacchi$^{1,7,8}$
  \space\space
  David B. Lindell$^{1,2}$
  \\
  \space\space
  \small{\textnormal{$^{1}$University of Toronto\space\space$^{2}$Vector Institute\space\space$^{3}$KAUST\space\space$^{4}$Snap Inc.\space\space$^{5}$Stanford University\space\space$^{6}$University of Michigan\space\space$^{7}$SFU\space\space$^{8}$Google}
}}

\begin{document}
\twocolumn[{
\maketitle
\vspace{-2em}
\begin{center}
    \centering
    \includegraphics[width=6.9in]{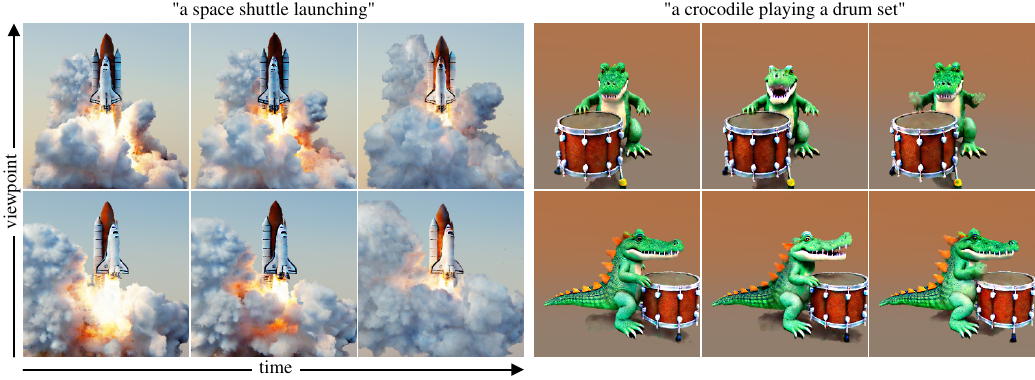}
    \captionof{figure}{\textbf{Text-to-4D Synthesis.} We present 4D-fy, a technique that synthesizes 4D (i.e., dynamic 3D) scenes from a text prompt. We show scenes generated from two text prompts for different viewpoints (vertical dimension) at different time steps (horizontal dimension). Video results can be viewed on our website: \url{https://sherwinbahmani.github.io/4dfy}.}
\label{fig:teaser}
\end{center}
}]
\maketitle
% Sections
\begin{abstract}
\vspace{-0.5em}
Recent breakthroughs in text-to-4D generation rely on pre-trained text-to-image and text-to-video models to generate dynamic 3D scenes. However, current text-to-4D methods face a three-way tradeoff between the quality of scene appearance, 3D structure, and motion. For example, text-to-image models and their 3D-aware variants are trained on internet-scale image datasets and can be used to produce scenes with realistic appearance and 3D structure---but no motion. Text-to-video models are trained on relatively smaller video datasets and can produce scenes with motion, but poorer appearance and 3D structure. While these models have complementary strengths, they also have opposing weaknesses, making it difficult to combine them in a way that alleviates this three-way tradeoff. Here, we introduce hybrid score distillation sampling, an alternating optimization procedure that blends supervision signals from multiple pre-trained diffusion models and incorporates benefits of each for high-fidelity text-to-4D generation.
Using hybrid SDS, we demonstrate synthesis of 4D scenes with compelling appearance, 3D structure, and motion.
\end{abstract}
    
\section{Introduction}
\label{sec:introduction}

The advent of internet-scale image--text datasets~\cite{schuhmann2022laion} and advances in diffusion models~\cite{ho2020denoising,song2020score,sohl2015deep} have led to new capabilities in stable, high-fidelity image generation from text prompts~\cite{saharia2022photorealistic,rombach2022high,balaji2022ediffi}.
Recent methods have also shown that large-scale text-to-image or text-to-video~\cite{singer2022make} diffusion models learn useful priors for 3D~\cite{jain2022zero,poole2022dreamfusion} and 4D scene generation~\cite{singer2023text}.  
Our work focuses on text-to-4D scene generation (Fig.~\ref{fig:teaser}), which promises exciting new capabilities for applications in augmented and virtual reality, computer animation, and industrial design.

Current techniques for generating 3D or 4D scenes from text prompts typically iteratively optimize a representation of the scene using supervisory signals from a diffusion model~\cite{wang2023prolificdreamer,poole2022dreamfusion, wang2023score}.
Specifically, these methods render an image of a 3D scene, add noise to the rendered image, use a pre-trained diffusion model to denoise the rendered image, and estimate gradients used to update the 3D representation~\cite{poole2022dreamfusion, wang2023score}.
This procedure, known as score distillation sampling (SDS)~\cite{poole2022dreamfusion}, underpins most recent methods for text-conditioned scene generation.

\begin{table}
\caption{Text-to-4D models face a tradeoff between the quality of appearance, 3D structure, and motion depending on the type of generative model used for score distillation sampling (SDS): text-to-image (T2I), 3D-aware T2I, or, text-to-video (T2V).} 
\small
\centering
\resizebox{\columnwidth}{!}{%
\begin{tabular}{lccc}
    \toprule
    \textit{SDS model} & \textit{appearance} & \textit{3D structure} & \textit{motion}\\ \midrule
T2I~\cite{saharia2022photorealistic,rombach2022high,yu2023scaling,balaji2022ediffi}  & \cellcolor{tabgreen} high & \cellcolor{tabred} low & \cellcolor{tabred} N/A  \\
3D-aware T2I~\cite{shi2023mvdream,li2023sweetdreamer}      & \cellcolor{tabyellow} medium   & \cellcolor{tabgreen} high   & \cellcolor{tabred} N/A \\ 
T2V~\cite{singer2022make,ho2022imagen,blattmann2023align,wang2023videofactory,wu2023lamp}  & \cellcolor{tabred} low   & \cellcolor{tabred} low   & \cellcolor{tabgreen} high  \\ \midrule
\textbf{Our method}  & \cellcolor{tabyellow} medium   & \cellcolor{tabgreen} high   & \cellcolor{tabyellow} medium  \\\bottomrule
\end{tabular}
}
\label{tab:tradeoff}
\end{table}

\begin{figure}[t]
\centering
\includegraphics[width=3.3in]{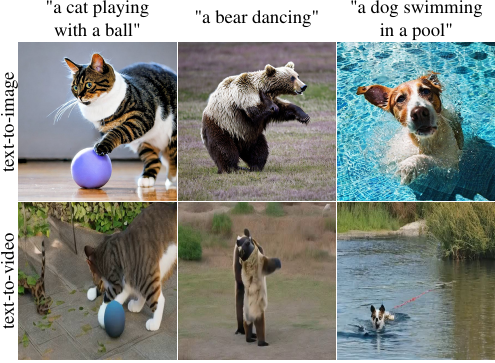}
\captionof{figure}{\textbf{Comparing text-to-image and text-to-video models.} Rendered frames from Stable Diffusion version 2.1 (top; text-to-image) ~\cite{stablediffusion} and Zeroscope version 2 (bottom; text-to-video) ~\cite{zeroscope} show significant disparity in appearance, with the text-to-image model appearing far more realistic.}
\label{fig:2d_sampling}
\vskip -0.1in
\end{figure}

Using SDS for text-to-4D generation requires navigating a three-way tradeoff between the quality of appearance, 3D structure, and motion (see Table~\ref{tab:tradeoff}); existing techniques obtain satisfactory results in just one or two of these categories.
For example, while SDS produces images that \textit{appear} realistic when rendering a generated scene from any particular viewpoint, inspecting multiple viewpoints can reveal that the scene has several faces or heads, replicated appendages, or other incorrectly repeated 3D structures---an issue now referred to as the ``Janus problem''~\cite{shi2023mvdream}.\footnote{Referring to the two-faced Roman god of beginnings and endings.} 
One way to improve 3D structure is to use SDS with a 3D-aware diffusion model that is trained to generate images from different camera viewpoints~\cite{liu2023zero}. 
But 3D-aware models sacrifice appearance quality as they require fine-tuning on synthetic datasets of posed images~\cite{shi2023mvdream}.
Incorporating motion into a scene using SDS with a text-to-video model~\cite{wang2023videofactory} typically degrades the appearance relative to static scenes generated with text-to-image models, which are more realistic (see Fig.~\ref{fig:2d_sampling}).
While different types of diffusion models thus have complementary qualities, they also have opposing weaknesses (Table~\ref{tab:tradeoff}).
Therefore, it is not trivial to combine them in a way that yields text-to-4D generation with high-quality appearance, 3D structure, and motion. 

Here, we propose a method for text-to-4D scene generation that alleviates this three-way tradeoff using \textit{hybrid~SDS}, an alternating optimization scheme that blends gradient updates from multiple pre-trained diffusion models and synthesizes 4D scenes using the best qualities of each. The method consists of three stages of optimization:~(1) we use a 3D-aware text-to-image model~\cite{shi2023mvdream} to generate an initial static 3D scene (without the Janus problem);~(2) we continue the optimization by blending in alternating supervision with variational SDS~\cite{wang2023prolificdreamer} and a text-to-image model to improve appearance;~(3) we blend in alternating supervision using video SDS with a text-to-video model~\cite{wang2023videofactory} to add motion to the scene.
By smoothly incorporating supervisory signals from these three diffusion models throughout the training process, we achieve text-driven 4D scene generation with state-of-the-art quality in terms of appearance, 3D structure, and motion. 
Overall we provide the following contributions.
\begin{itemize}
    \item We introduce hybrid SDS, a technique that extracts desirable qualities from multiple pre-trained diffusion models and alleviates a tradeoff between appearance, 3D structure, and motion in text-to-4D scene generation.
    \item We provide a quantitative and qualitative evaluation of the method, and we explore the three-way tradeoff space with ablation studies to facilitate future research.
    \item We demonstrate text-to-4D generation based on open-source pretrained models and will make all codes and evaluation procedures publicly available.
    \item We present state-of-the-art results for the task of text-to-4D generation.
\end{itemize}

\section{Related Work}
\label{sec:related}

Our method is related to techniques from multiple areas of generative modeling, including text-to-image, text-to-video, and text-to-3D models. For more extensive discussions of related works, we refer readers to a recent state-of-the-art report on diffusion models~\cite{po2023state}.

\paragraph{Text-to-image generation.}
Methods for generating images from text prompts are a relatively new innovation, first demonstrated using generative adversarial networks~\cite{reed2016generative,xu2018attngan,zhang2017stackgan}. 
The problem itself is also related to other methods for text-based image retrieval~\cite{liu2007survey} or image-conditioned text generation~\cite{you2016image,stefanini2022show}.
More recently, models trained on text--image datasets with billions of samples~\cite{schuhmann2022laion} have become the state of the art for this task~\cite{rombach2022high}.

Diffusion models~\cite{ho2020denoising, song2020denoising} are a popular architecture for generative modeling on large-scale datasets, and autoregressive models have also shown promising results~\cite{yu2022scaling,ramesh2021zero}. 
Typically, these methods exploit a pretrained text encoder, such as CLIP~\cite{radford2021learning}, to encode the text prompt into a feature vector used to condition the diffusion model~\cite{nichol2021glide,ramesh2022hierarchical}.
In diffusion models, high-resolution (i.e.,~megapixel) image generation is achieved by applying repeated upsampling layers~\cite{ramesh2022hierarchical, ho2022cascaded} or performing diffusion in the lower-resolution latent space of an autoencoder and then decoding the result to recover an image at the nominal resolution~\cite{gu2022vector, rombach2022high}.
Our work incorporates two open-source text-to-image diffusion models: Stable Diffusion~\cite{rombach2022high} and MVDream~\cite{shi2023mvdream} (a recent 3D-aware diffusion model) to enable 4D scene generation.

\paragraph{Text-to-video generation.}
Our work relies on the burgeoning field of video generation via diffusion models, an area that is somewhat constrained by the limited scale of video datasets.
To counteract this, methods often utilize a hybrid training approach on both image and video datasets, such as WebVid-10M~\cite{bain2021frozen}, HD-VG-130M~\cite{wang2023videofactory}, or HD-VILA-100M~\cite{xue2022advancing}.
Recent approaches in this field typically employ variations of pixel-space upsampling (both in space and time)~\cite{ho2022imagen} or latent space upsampling to improve spatial and temporal resolution~\cite{guo2023animatediff,he2022latent,wang2023videocomposer,zhou2022magicvideo}.
Autoregressive models distinguish themselves by their ability to generate videos of varying lengths~\cite{villegas2022phenaki}. 
Further improvements in video synthesis have been achieved by finetuning pre-trained text-to-image diffusion models on video data~\cite{blattmann2023align,singer2022make,wu2023lamp}, or separating the content and motion generation process by using an initial image frame as a starting point \cite{wu2023lamp, guo2023animatediff}.
Despite recent advances in text-to-video synthesis, the fidelity of generated videos still lags behind that of static image generation (see Fig.~\ref{fig:2d_sampling}) and so they perform poorly when used directly with SDS for text-to-4D generation. 
Instead, our work leverages an open-source latent space text-to-video diffusion model called Zeroscope~\cite{zeroscope} (extended from the Modelscope architecture~\cite{wang2023modelscope}) together with other pre-trained, open-source diffusion models using hybrid SDS.

\paragraph{Text-to-3D generation.}
Early methods for text-to-3D generation relied on parsers to convert input text to a semantic representation and synthesized scenes from an object database~\cite{adorni1983natural,coyne2001wordseye,chang2014learning}.
Later, automated, data-driven methods used multi-modal datasets~\cite{chen2018text2shape}, and pre-trained models, such as CLIP~\cite{radford2021learning}, to edit or stylize an input 3D mesh~\cite{jetchev2021clipmatrix, gao2023textdeformer} or a radiance field~\cite{wang2022clip}. 
More recently, CLIP-based supervision enabled synthesis of entire 3D scenes~\cite{jain2022zero, sanghi2022clip}, and these techniques evolved into the most recent approaches, which optimize a mesh or radiance field based on SDS supervision~\cite{poole2022dreamfusion, wang2023prolificdreamer, lin2022magic3d}.
The quality of their 3D structures has been improved by applying diffusion models that consider multiple viewpoints~\cite{lin2023consistent123, liu2023zero, shi2023mvdream}. 
Alternatively, recent advancements have seen a  shift towards using diffusion or transformer models to transform an input 2D image into a  3D representation for novel-view synthesis \cite{chan2023generative, tang2023make, gu2023nerfdiff, liu2023syncdreamer, yoo2023dreamsparse, tewari2023diffusion, qian2023magic123}. 
Still, these techniques do not yet support generating 4D scenes.

Our work is most closely related to Make-A-Video3D~(MAV3D)~\cite{singer2023text}, a recent method for text-to-4D generation that integrates SDS-based supervision in two separate stages: first with a text-to-image model and subsequently with a text-to-video model.
Similar to MAV3D, we aim to generate dynamic 3D scenes; however, our approach uses hybrid SDS, which allows gradient updates from multiple models to be smoothly blended together in an alternating optimization. 
Our approach generates high-quality dynamic 3D scenes and does not suffer from Janus problems.

\paragraph{Concurrent works.}
Concurrent works on text-to-4D~\cite{ling2023align,zheng2023unified}, image-to-4D~\cite{ren2023dreamgaussian4d,zheng2023unified,zhao2023animate124}, and video-to-4D~\cite{jiang2023consistent4d, yin20234dgen, pan2024fast} similarly use recent diffusion models for 4D generation. %We believe that hybrid SDS is orthogonal to these approaches.

\begin{figure*}[t]
    \begin{center}
\includegraphics[width=6.9in]{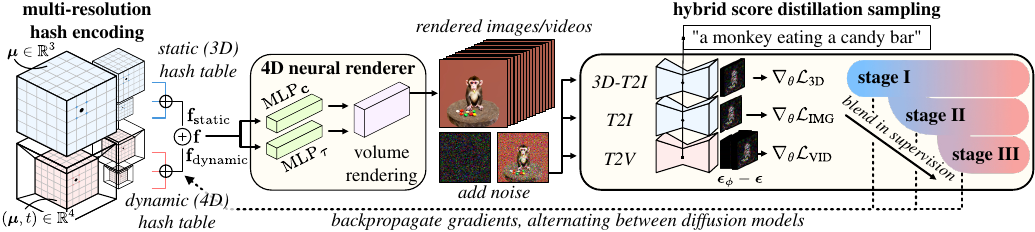}
\end{center}
\caption{\textbf{Overview}.
A 4D radiance field is parameterized using a neural representation with a static and dynamic multiscale hash table of features. Images and videos are rendered from the representation using volume rendering, and we supervise the representation using hybrid score distillation sampling---a technique that combines gradients from multiple types of pre-trained diffusion models. 
In the first stage of training we use gradients $\msds$ from a 3D-aware text-to-image model~(3D-T2I) to iteratively optimize a representation without the~Janus problem.
Next, we blend in gradient supervision using variational SDS with a text-to-image model~(T2I) to improve the appearance~(i.e., we alternate supervision between $\vsd$ and $\msds$).
In the last stage we incorporate gradients ($\vsds$) from a text-to-video model~(T2V) to add motion to the scene, and we update the scene using the other models in an alternating fashion.}
\vspace{-0.5em}
\label{fig:method_main}
\end{figure*}

\section{Method}
\label{sec:method}
   
Our approach for text-to-4D generation builds upon a hash-encoding-based neural representation~\cite{mueller2022instant} that implicitly decomposes the scene into static and dynamic feature grids~\cite{turki2023suds}.
In this section we overview our representation for 4D neural rendering and describe the optimization procedure based on hybrid SDS (see Fig.~\ref{fig:method_main}).

\subsection{4D Neural Rendering}
    
Volumetric neural rendering methods represent a scene using a neural representation to parameterize the attenuation and emission of light at every point in 3D space~\cite{mildenhall2021nerf,lombardi2019neural}.
We can use such a representation to render an image by casting a ray from the camera center of projection, through each pixel location, and into the scene. 
For sampled points along the ray $\raysamp \in\mathbb{R}^3$, we query a neural representation to retrieve a volumetric density $\densityval\in\mathbb{R}_+$ and color $\colorval\in\mathbb{R}_+^3$, which describe attenuation and emission of light, respectively, at a particular point.
Then, the resulting density and color samples are alpha-composited to recover the color of a rendered pixel $\pixelcolor$ as 
\begin{equation}
    \pixelcolor = \sum_i \weightval_i\colorval_i, \ \weightval_i= \alphaval\prod_{j<i}(1 - \alphaval_j),
    \label{eq:volume-rendering}
\end{equation}
where $\alphaval_i = 1 - e^{-\densityval_i \lVert \raysamp_i - \raysamp_{i+1}  \rVert}$.
We query the neural representation using an additional input time variable $\timeval$, which enables modeling time-varying density and color.

We illustrate the neural representation in Fig.~\ref{fig:method_main}; it consists of two multi-resolution hash tables to disentangle static and dynamic scene modeling.

Following Müller et al.~\cite{mueller2022instant}, the static hash table stores learnable feature vectors that are indexed by a voxel-lookup and hashing operation and decoded into density and color using two small multilayer perceptrons (MLPs).    
Concretely, we consider the neural representation 
\begin{equation}
\network: \raysamp, \timeval \rightarrow \densityval, \colorval
\label{eq:neural-representation}
\end{equation}
with $\params = \{\params_\text{static}, \params_\text{dynamic}, \theta_{\textsc{mlp}} \}$ denoting all learnable parameters from the static and dynamic hash tables and the MLPs.
For a given $\raysamp$, we query the static hash table by identifying the closest voxel at each scale $1{\leq}s{\leq}S$.
Then, we trilinearly interpolate the feature values from the voxel vertices after retrieving them from the hash table.
Retrieved features from each scale are concatenated as $\feature_\text{static} = \feature_\text{static}^{(1)}\oplus\cdots\oplus\feature_\text{static}^{(S)}$.
We follow the same procedure to query the dynamic hash table given ($\raysamp$, $\timeval$), except we use quadrilinear interpolation to interpolate feature values.
The resulting features from the static and dynamic hash tables are added as $\feature = \feature_\text{static} + \feature_\text{dynamic}$.
We do not model view-dependent effects in the feature encoding.
Finally, we decode density and color as $\textsc{mlp}_\densityval(\feature)$ and $\textsc{mlp}_\colorval(\feature)$, respectively. 

\subsection{Hybrid Score Distillation Sampling}

We leverage the 4D representation along with SDS to create dynamic 3D scenes from a text prompt.
Our hybrid approach incorporates three different flavors of SDS that are smoothly merged during an alternating optimization procedure to improve the structure and quality of the 4D model:
\begin{enumerate}
    \item SDS applied to a 3D-aware text-to-image diffusion model to optimize a static scene without the Janus problem.
    \item Variational score distillation sampling (VSD; a modified version of SDS~\cite{wang2023prolificdreamer}) using a standard text-to-image model~\cite{rombach2022high} to improve the appearance of the static scene.
    \item Video SDS using a text-to-video model~\cite{wang2023videofactory}, which extends SDS to multiple video frames and adds motion to the scene. 
\end{enumerate}
In the following, we describe each type of SDS and how it used for text-to-4D generation.

\paragraph{3D-aware scene optimization.} 
We first consider optimizing a static scene using SDS with a 3D-aware text-to-image diffusion model~\cite{shi2023mvdream}.
The diffusion model is pre-trained using a stochastic forward process that slowly adds Gaussian noise to multiview images $\mathbf{x}$ over timesteps $0{\leq} \timediff{\leq} T_d$.
With increasing $\timediff$, the process yields noisy images $\noise{\timediff}$ that, at $\timediff=T_d$, are close to zero-mean Gaussian. 
After training, the model reverses this process to add structure to the noisy images.
It predicts $\predimg(\noise{\timediff}; \timediff, \textcond, \extrinsics)$, which approximates the output of an optimal denoiser at each timestep $\timediff$, conditioned on a text embedding $\textcond$~\cite{saharia2022photorealistic, ramesh2022hierarchical, rombach2022high} and the camera extrinsics $\extrinsics$ corresponding to each image. 
In practice, text-to-image diffusion models typically predict the noise content $\prednoise$ rather than the denoised image $\predimg$. 
But note that the denoised image can still be obtained as $\predimg(\noise{\timediff}; \timediff, \textcond, \extrinsics)\propto \noise{\timediff}-\prednoise(\noise{\timediff}; \timediff, \textcond, \extrinsics)$, i.e., by subtracting the predicted noise from the noisy image~\cite{ho2020denoising}.
We implement 3D-aware SDS by rendering multiple images $\gtimg_\params$ from the neural representation, adding noise $\gtnoise$, and using the 3D-aware diffusion model~\cite{shi2023mvdream} to predict the noise~$\prednoise$ using classifier-free guidance~\cite{ho2022classifier}.
To update the parameters $\params$ of the neural representation, we use the 3D-aware SDS gradient: 
\begin{equation}
    \nabla_\params \mathcal{L}_{\textsds{}} = \mathbb{E}_{\timediff, \gtnoise, \extrinsics}\left[\weighting\left(\prednoise(\noise{\timediff}; \timediff, \textcond, \extrinsics) - \gtnoise\right) \frac{\partial \gtimg_\params}{\partial\params}\right],
    \label{eq:m-sds}
\end{equation}
where $\weighting$ is a weighting function that depends on the diffusion timestep, and we add a stop gradient to the output of the diffusion model~\cite{shi2023mvdream}.
Intuitively, the SDS loss queries the diffusion model to see how it adds structure to an image, then this information is used to backpropagate gradients to the scene representation.

\paragraph{Improving appearance using VSD.} 
We incorporate an additional loss term based on VSD~\cite{wang2023prolificdreamer} to improve the appearance of images rendered from the scene. 
This term uses a pre-trained text-to-image model~\cite{rombach2022high} along with a finetuning scheme that improves image quality over the 3D-aware text-to-image model alone. 
We follow Wang et al.~\cite{wang2023prolificdreamer} and augment the standard SDS gradient with the output of an additional text-to-image diffusion model that is finetuned using a low-rank adaptation~\cite{hu2021lora}, during scene optimization. 
Specifically, we have
\begin{equation}
    \resizebox{\columnwidth}{!}{
    $\nabla_\params \mathcal{L}_{\textvsd{}} = \mathbb{E}_{\timediff, \gtnoise, \extrinsics}\left[\weighting\left(\prednoise(\noise{\timediff}; \timediff, \textcond) - \prednoisefinetune(\noise{\timediff}; \timediff, \textcond, \extrinsics)\right) \frac{\partial \gtimg_\params}{\partial\params}\right]$},
    \label{eq:vsd}
\end{equation}
where $\prednoisefinetune$ is the noise predicted using a finetuned version of the diffusion model that incorporates additional conditioning from the camera extrinsics $\extrinsics$; here, we let $\noise{\timediff}$ represent a noisy version of a single image rendered from $\network$.
The model is finetuned using the standard diffusion objective
\begin{equation}
    \min_\theta \ \mathbb{E}_{\timediff,\gtnoise,\extrinsics}\left[\lVert \prednoisefinetune(\noise{\timediff}; \timediff, \textcond, \extrinsics) - \gtnoise \rVert_2^2 \right]. 
    \label{eq:finetune}
\end{equation}
Note that, different from the original description of~VSD~\cite{wang2023prolificdreamer}, we find we can omit the simultaneous optimization over multiple scene samples {(i.e. the variational component of~\cite{wang2023prolificdreamer}), which reduces memory requirements without significantly degrading appearance.

\paragraph{Adding motion with Video SDS.} 
Last, we use supervision from a text-to-video diffusion model~\cite{wang2023videofactory} to add motion to the generated scene.
This procedure extends the original~SDS gradient by incorporating structure added by the diffusion model to all noisy video frames~\cite{singer2023text}.
The video~SDS gradient is given as 
\begin{equation}
    \nabla_\params \mathcal{L}_{\text{\textvsds{}}} = \mathbb{E}_{\timediff, \gtnoise}\left[\weighting\left(\prednoise(\noise{\timediff}; \timediff, \textcond) - \gtnoise\right) \frac{\partial \gtvideo_\params}{\partial\params}\right].
    \label{eq:v-sds}
\end{equation}
To simplify notation, we re-use $\prednoise$ and $\gtnoise$ to here represent the predicted and actual noise for each video frame, and we let $\gtvideo_\params$ be a collection of $V$ video frames $\gtvideo_\params = [\gtimg_\params^{(1)}, \ldots, \gtimg_\params^{(V)}]^T$ rendered from the representation.

\begin{algorithm}[t]
\caption{Hybrid Score Distillation Sampling}\label{alg:cap}
\label{alg:hybrid}
\small
\textbf{Require:} \\
\hspace*{\algorithmicindent}   $\network$ \Comment{4D neural representation}\\
\hspace*{\algorithmicindent}   $\itersa$, $\itersb$, $\itersc$ \Comment{iterations for each stage}\\
\hspace*{\algorithmicindent}   $\parammsds$, $\paramvsd$ \Comment{update probabilities}\\
\hspace*{\algorithmicindent}   $\nabla_\params \mathcal{L}_{\textsds{}}$, $\nabla_\params \mathcal{L}_{\textvsd{}}$, $\nabla_\params \mathcal{L}_{\textvsds{}}$ \Comment{SDS grads. (Eqs.~\ref{eq:m-sds},~\ref{eq:vsd},~\ref{eq:v-sds})}

\begin{algorithmic}[1]
   \vspace{1em}
    \State // \textbf{Stage 1}
    \State freeze dynamic hash map ($\params_\text{dynamic}$)
    \For{iter in $\itersa$} \Comment{\textsds{} update}
        \State $\mathbf{grad}=
        \begin{cases}
            \msds\\ 
        \end{cases}$
        \State $\Call{Update}{\mathbf{grad}}$
    \EndFor

    \vspace{1em}
    \State // \textbf{Stage 2}
    \For{iter in $\itersb$} \Comment{\textsds{} or \textvsd{} update}
        \State $\mathbf{grad}=
        \begin{cases}
            \msds, & \text{\hspace{-0.75em}with probability } \parammsds\\ 
            \vsd, & \text{\hspace{-0.75em}otherwise}\\
        \end{cases}$
        \State $\Call{Update}{\mathbf{grad}}$

    \EndFor

    \vspace{1em}
    \State // \textbf{Stage 3}
    \State decrease learning rate of static hash map ($\params_\text{static}$)
    \For{iter in $\itersc$}\Comment{\textsds{}, \textvsd{}, or \textvsds{} update}
        \State $\mathbf{grad}=
        \begin{cases}
            \msds, & \text{\hspace{-0.75em}with probability }\parammsds\\ 
            \vsd, & \text{\hspace{-0.75em}with probability }\parammsds\cdot\paramvsd\\
            \vsds, & \text{\hspace{-0.75em}otherwise}
        \end{cases}$
        \State \textbf{if} \textvsds{}, unfreeze $\params_\text{dynamic}$
        \State $\Call{Update}{\mathbf{grad}}$

    \EndFor

    \vspace{1em}
    \Procedure{Update}{$\mathbf{grad}$}
        \State $\gtimg \leftarrow \network$ \Comment{render images (Eq.~\ref{eq:volume-rendering})}
        \State take gradient step on $\mathbf{grad}$ \Comment{optimize $\network$}
        \State \textbf{if} \textvsd{}, take finetuning step (Eq.~\ref{eq:finetune})
    \EndProcedure
\end{algorithmic}
\vspace{-0.5em}
\end{algorithm}

\paragraph{Optimization procedure -- Algorithm~\ref{alg:hybrid}.}
We optimize the 4D representation in three stages that smoothly blend supervision in alternating steps from~(1)~3D-aware SDS,~(2)~VSD, and~(3)~video SDS. 
\\ \indent \textbf{Stage 1.} In the first stage of optimization, we update $\network$ using gradients from 3D-aware SDS until convergence.
Since this stage focuses on optimizing a static scene, we freeze (i.e.~do not update) the parameters of the dynamic hash table $\feature_\text{dynamic}$ and only update the static hash table and decoder~MLP. 
We set the total number of first-stage iterations $\itersa$ to match that of~\citet{shi2023mvdream}, which allows the optimization to proceed until there are no distinguishable changes in the rendered scene from one iteration to the next. 
% \item
\\ \indent \textbf{Stage 2.} Next, we add VSD gradients using an alternating optimization procedure.
At each iteration, we randomly select to update the model using $\msds$ or $\vsd$ with probability $\parammsds$ and $\paramvsd$.
We continue this alternating optimization for $\itersb$ iterations, until convergence.
As we show in the next section, this stage of optimization results in improved appearance compared to using $\msds$ alone while also being free of the Janus problem. 
\\ \indent \textbf{Stage 3.}
Last, we update the representation using a combination of all gradient updates. 
Specifically, we randomly select to update the model at each iteration using $\msds$, $\vsd$, or $\vsds$ with probability $\parammsds$, $\parammsds\cdot\paramvsd$, and $1-\parammsds\cdot\paramvsd$, respectively.
Since we now aim to incorporate motion into the representation, we unfreeze the parameters of the dynamic hash table during the update with $\vsds$ but keep them frozen for updates using the text-to-image models.
We also decrease the learning rate of the static hash table to preserve the high-quality appearance from the previous stage.
We repeat the alternating optimization in the final stage until convergence, which we find occurs consistently within~$\itersc$ iterations.
Overall, hybrid SDS effectively combines the strengths of each pre-trained diffusion model while avoiding quality degradations that result from naively combining gradients from each model.

\begin{figure*}[t]
\centering
\includegraphics[width=6.9in]{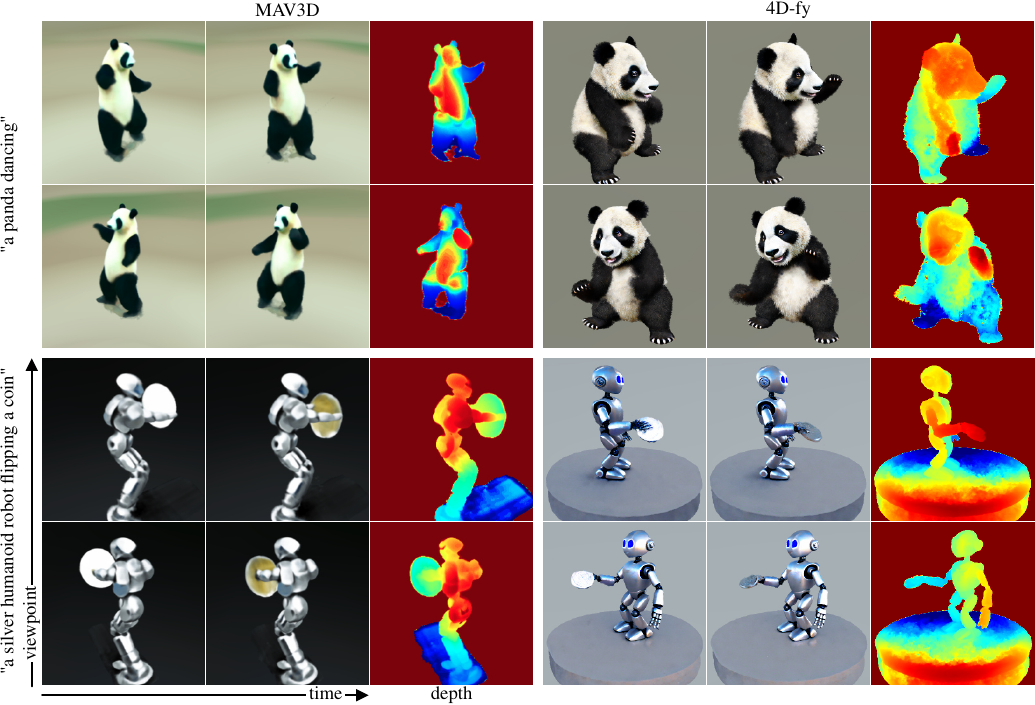}
\captionof{figure}{\textbf{Text-to-4D Comparison.}
We compare against MAV3D~\cite{singer2023text}, and observe our approach obtains significantly higher quality results.}
\vspace{-0.5em}
\label{fig:sota}
\end{figure*}

\subsection{Implementation}
We implement hybrid SDS based on the threestudio framework~\cite{threestudio}, which includes implementations of MVDream~\cite{shi2023mvdream}~(for 3D-aware text-to-image diffusion and SDS), ProlificDreamer~\cite{wang2023prolificdreamer} with Stable Diffusion~\cite{rombach2022high}~(text-to-image diffusion and VSD), and we implement the video SDS updates using Zeroscope~\cite{wang2023videofactory,zeroscope}.

\paragraph{Hyperparameter values.}
We initialize the 4D neural representation following \cite{poole2022dreamfusion, lin2022magic3d} and add an offset to the density predicted by the network in the center of the scene to promote object-centric reconstructions. 
We set the learning rates for the static hash map to 0.01, for the dynamic hash map to 0.01, and for the MLP to 0.001.
We drop the learning rate for the static hash map to 0.0001 before the last stage to focus the gradient updates on the dynamic hash map.
The values of $\itersa$, $\itersb$, and $\itersc$ are set to 10000, 10000, and 100000, respectively.
We set the probabilities for hybrid SDS to $\parammsds=0.5$ and $\paramvsd=0.5$ for a reasonable tradeoff with respect to appearance, 3D structure, and motion.

\paragraph{Rendering.}
Each of the diffusion models has a different native resolution, so we render images from $\network$ accordingly.
We render four images from different camera positions for the 3D-aware SDS at the native (256$\times$256 pixel) resolution of the 3D-aware text-to-image model.
The VSD update is computed by rendering a 256$\times$256 image and bilinearly upsampling the image to the native resolution of Stable Diffusion (512$\times$512).
Finally, the video SDS update is computed by rendering 16 video frames at 160$\times$288 resolution and upsampling to the native 320$\times$576 resolution of Zeroscope. 
%To render a background for the image, we optimize a second small MLP that takes in a ray direction and returns a background color. 
%We composite the ray color $\pixelcolor$ rendered from $\network$ on top of this background color.

%\paragraph{Computation.}
%We optimized the model on an NVIDIA A100 GPU. 
%The entire procedure requires roughly 80 GB of VRAM and the three optimization stages require approximately 2, 2, and 19 hours of compute, respectively.

\section{Experiments}
\label{sec:experiments}

\begin{table}[!t]
    \caption{\textbf{Quantitative results.} We compare our method against MAV3D and variations of 4D-fy with different loss terms or backbone architectures (i.e., with HexPlane~\cite{cao2023hexplane}). The methods are evaluated in terms of CLIP Score (CLIP) and human preference based on appearance quality (AQ), 3D structure quality (SQ), motion quality (MQ), text alignment (TA), and overall preference (Overall). The numbers reported for human preference are the percentages of users who voted for our method over the corresponding method in head-to-head comparisons.}
    \label{tab:results}
    \begin{center}
    \resizebox{\columnwidth}{!}{
    \begin{tabular}{lc|cccc|c}
        \toprule
         &  & \multicolumn{5}{c}{\textit{Human Preference}}\\
        \textit{Method} & \textit{CLIP} & AQ & SQ & MQ & TA & Overall \\\midrule
        MAV3D~\cite{singer2023text} & 33.9 & 92\% & 89\% & 41\% & 52\% & 67\%\\
        \textbf{4D-fy} & 34.2 & \multicolumn{4}{c|}{---} & ---\\
        \midrule
        \textit{Ablation Study} & \multicolumn{6}{c}{}  \\\midrule
        \textbf{4D-fy} & 35.0 & \multicolumn{4}{c|}{---} & ---\\
        w/o $\nabla_\theta\mathcal{L}_\text{\textsds{}/\textvsd{}}$ & 29.3 & 100\% & 100\% & 78\% & 86\% & 94\% \\
        w/o $\msds$ & 35.1 & 88\% & 89\% & 95\% & 92\% & 91\% \\
        w/o $\vsd$ & 34.5 & 70\% & 68\% & 68\% & 69\% & 70\% \\
        w/o hybrid SDS & 33.8 & 100\% & 100\% & 78\% & 88\% & 95\%\\%\midrule
        w/ HexPlane & 34.5 & 95\% & 92\% & 90\% & 92\% & 95\% \\
        \bottomrule
    \end{tabular}}
    \end{center}
    \vskip -0.2in
\end{table}

\subsection{Metrics}
\label{sec:metrics}

We assess our method using CLIP Score~\cite{park2021benchmark} and a user study. We compare our model against MAV3D for 28 prompts and against our ablations for a subset of 5 prompts. Current text-to-4D models are costly to train, and many researchers in academia do not have access to the scale of resources available to large tech companies. Hence, we only used a subset due to computational limitations. To promote future research in this field, we open source the evaluation protocol for the user study along the code: \url{https://github.com/sherwinbahmani/4dfy}.

\paragraph{CLIP Score.} CLIP Score~\cite{park2021benchmark} evaluates the correlation between a text prompt and an image. Specifically, this corresponds to the cosine similarity between textual CLIP \cite{radford2021learning} embedding and visual CLIP \cite{radford2021learning} embedding. The score is bound between 0 and 100, where 100 is best. We calculate the CLIP score for MAV3D using the same procedure we use for our method. Specifically, for each input text prompt, we render a video using the same camera trajectory as MAV3D, i.e., moving around the scene in azimuth with a fixed elevation angle. Subsequently, we score each video frame with~CLIP ViT-B/32 and average the scores over all frames and text prompts to derive the final CLIP score.

\paragraph{User study.}
We conduct qualitative comparisons between our method and the baseline, MAV3D, by surveying 26 human evaluators. We use the same head-to-head comparison model as the user survey conducted by MAV3D. Specifically, we present text prompts alongside the corresponding outputs of our method and the baseline method in random order. Evaluators are requested to specify their overall preference for a video, as well as evaluate four specific properties: appearance quality, 3D structure quality, motion quality, and text alignment. In Table~\ref{tab:results}, we report the percentage of users who prefer each method overall and based on each of the four properties. We conduct $\chi^2$-tests to evaluate statistical significance at the $p < 0.05$ level. Further details on the user study are included in the supplementary.

\subsection{Results}
\label{sec:results}

We visualize spatio-temporal renderings along with depth maps in comparison to MAV3D in Fig.~\ref{fig:sota}. Although both methods can synthesize 4D scenes, MAV3D noticeably lacks detail. In contrast, our method produces realistic renderings across space and time. 
We report quantitative metrics in Table~\ref{tab:results}. 
In terms of CLIP Score and overall preference in the user study 4D-fy outperforms MAV3D. Users indicated a statistically significant preference towards 4D-fy compared to MAV3D in terms of appearance quality, 3D structure quality, text alignment, and overall preference. They rated the motion quality roughly on par with MAV3D, which used a propriety text-to-video model. For example, overall, 67\% of users prefer our method over 33\% for MAV3D.

\begin{figure*}[t]
\centering
\includegraphics[width=6.9in]{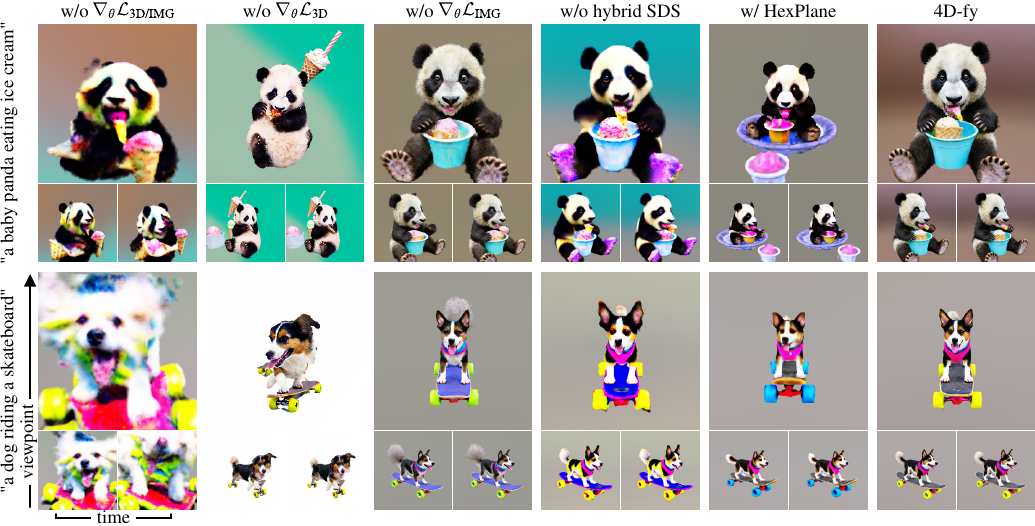}
\captionof{figure}{\textbf{Ablation study.} We assess the qualitative impact of removing gradient updates from different models during optimization. Our method without image guidance ($\nabla_\theta\mathcal{L}_\text{\textsds{}/\textvsd{}}$) does not produce realistic appearance and 3D structure. Removing the 3D-aware guidance ($\msds$) generates high-quality appearance but low-quality 3D structure. Our approach without~VSD~($\vsd$) reduces the appearance quality. Hybrid SDS is crucial for appearance and 3D structure, while using HexPlane reduces the appearance quality. Best viewed digitally.}
\label{fig:ablation}
\end{figure*}

\subsection{Ablations}
\label{sec:ablations}
We provide an in-depth analysis motivating our hybrid SDS training scheme by ablating each component and evaluating the use of a 4D neural representation more similar to that of MAV3D.
We provide ablations in Table~\ref{tab:results} and in Fig.~\ref{fig:ablation}.

\paragraph{Image guidance (w/o $\nabla_\theta\mathcal{L}_\text{\textsds{}/\textvsd{}}$).}
Technically, learning a dynamic 3D scene solely from a text-to-video model without text-to-image guidance is possible.
To demonstrate the drawbacks of this approach, we present results where we skip the first two stages and directly train the model with text-to-video guidance only.
This corresponds to setting $\parammsds=0$ and $\paramvsd=0$.
Our experiments reveal that the text-to-video model fails to provide realistic 3D structure and high-quality appearance for generating a dynamic 3D scene.

\paragraph{3D-aware guidance (w/o $\msds$).}
We find that using a 3D-aware diffusion model is crucial for generating realistic 3D structures.
If we remove the 3D-aware diffusion model, i.e., by setting $\parammsds=0$, we can generate scenes with similar motion and high-quality appearance, but the 3D structure is degraded. This is evident for both scenes in Fig.~\ref{fig:ablation}.

\paragraph{VSD guidance (w/o $\vsd$).}
We find that VSD helps provide a realistic scene appearance; if we disable it during scene generation, i.e., $\paramvsd=0$, there are some negative effects.
For example in Fig.~\ref{fig:ablation}, the ice cream cone in the bucket (top row) is more detailed, and the dog's face (bottom row) is sharper (please zoom in). 

\paragraph{Hybrid SDS.}
To illustrate the impact of our hybrid SDS approach we disable image guidance after the second stage by setting $\parammsds=0$ and $\paramvsd=0$ for the third stage only. 
This aligns with the MAV3D training scheme, where a static model is pre-trained with text-to-image and subsequently fine-tuned with text-to-video.
Our quantitative and qualitative analysis shows that this approach results 
in degraded appearance and 3D structure.
We find that incorporating text-to-image, 3D-aware text-to-image, and text-to-video via hybrid SDS in the final optimization stage preserves a realistic appearance and high-quality 3D structure.

\paragraph{Backbone architecture.}
Finally, we ablate the hash-grid-based 4D representation by replacing it with the HexPlane \cite{cao2023hexplane, fridovich2023k} architecture. 
This representation similarly disentangles static and dynamic scene components and can be readily integrated into our pipeline.
The HexPlane approach fails to match the appearance quality of the hash-grid-based representation.
MAV3D uses HexPlane but implements a multi-scale variant with a large 5-layer decoding MLP featuring 128 hidden units.
We could not re-implement this approach as the model does not fit on an 80 GB A100 GPU.
To allow for a fair comparison, we instead increased the capacity of HexPlane to match the memory consumption of our hash-grid-based representation.
We expect that increasing the capacity of HexPlane and longer training times could lead to similar results as our representation.
\section{Conclusion}
Our method synthesizes high-quality 4D scenes from text prompts using a novel hybrid score distillation sampling procedure.
Our work alleviates a three-way tradeoff between appearance, 3D structure, and motion and is the first to build on open-source models. 
We will release the code to facilitate future research in text-to-4D generation. 

\paragraph{Limitations.}
Although our method produces compelling dynamic 3D scenes, there are several limitations and avenues for future work. 
First, the complexity of motion in our scenes is limited to simple movements. 
We believe that our method will directly benefit from future progress in text-to-video generation, as current text-to-video models suffer from low-quality renderings and unrealistic motion. 
Another way to improve motion could be exploiting recently proposed dynamic representations, e.g., dynamic 3D Gaussians~\cite{luiten2023dynamic}.
Moreover, current metrics in text-to-3D generation are not sufficient, as they mainly rely on image-based metrics and user studies. 
Designing more sophisticated 3D and 4D metrics is an important direction for future work. 
Lastly, generating each scene takes a significant amount of time. Concurrent text-to-3D works~\cite{hong2023lrm, li2023instant3d} alleviate this problem by training a large-scale model on 3D data, allowing generation within seconds. 
Incorporating our hybrid optimization procedure to blend between large-scale pre-training on 2D, 3D, and video data could enable fast text-to-4D generation.

\paragraph{Ethics Statement.} We condemn the application of our method for creating realistic fake content intended to harm specific entities or propagate misinformation.
\section{Acknowledgements}

This work was supported by the Natural Sciences and Engineering Research Council of Canada (NSERC) Discovery Grant program, the Digital Research Alliance of Canada, and by the Advanced Research Computing at Simon Fraser University. It was also supported in part by ARL grant W911NF-21-2-0104, a Vannevar Bush Faculty Fellowship, a
gift from the Adobe Corporation, a PECASE by the ARO, NSF award
1839974, Stanford HAI, and a Samsung GRO.
%\clearpage
{
    \small
    \bibliographystyle{ieeenat_fullname}
    \bibliography{ref}
}

\renewcommand\thesection{S\arabic{section}}
\renewcommand\thefigure{S\arabic{figure}}
\setcounter{section}{0}
\clearpage
\section{Implementation Details}
\label{sec:supp_implementation}

Here, we provide additional implementation details.

\paragraph{Optimization}
Following \cite{wang2023prolificdreamer, shi2023mvdream}, during the first stage of the optimization, we anneal the sampled diffusion timesteps over 5000 iterations from $\timediff\in[0.02, 0.98]$ to a range of $\timediff\in[0.02, 0.5]$.
In subsequent stages, we keep the timestep sampling the same, except for the video SDS updates, where we sample $\timediff\in[0.02, 0.98]$. For the $\vsds$ updates, we set the total learning rate to 0.1, while $\msds$ and $\vsd$ use a learning rate of 1.0.
To render a background for the image, we optimize a second small MLP that takes in a ray direction and returns a background color. 
We composite the ray color $\pixelcolor$ rendered from $\network$ on top of this background color.

\paragraph{Model}
Our hash-grid-based representation \cite{mueller2022instant} uses default values \cite{threestudio} for both the static and dynamic components: 16 levels, 2 features per level, and a base resolution of 16.

\paragraph{Rendering}
We use NerfAcc \cite{li2022nerfacc} as our rendering pipeline without any changes to the default values in threestudio \cite{threestudio}.
During training and inference we use 512 samples per ray.
Following the MAV3D \cite{singer2023text} evaluation protocol, we sample 64 frames during inference after training our model with 16 frames, which is possible due to the implicit time coordinate. During training, we sample a random time offset with evenly spaced time coordinates over the [0.0, 1.0] time interval. This enables smooth time sampling across the whole range of time coordinates.

\paragraph{Computation.}
We optimized the model on an NVIDIA A100 GPU. 
The entire procedure requires roughly 80 GB of VRAM and the three optimization stages require approximately 2, 2, and 19 hours of compute, respectively.

\paragraph{Runtime.}
Rendering one frame for our method takes 71ms vs. 68 ms for a static 3D MVDream based model. We cannot compare to MAV3D as code is not available, and they do not report runtimes.
\section{User Study}
\label{sec:supp_user_study}

\begin{figure}[t]
\centering
\includegraphics[trim={4cm 12cm 4cm 2cm},clip,width=\columnwidth]{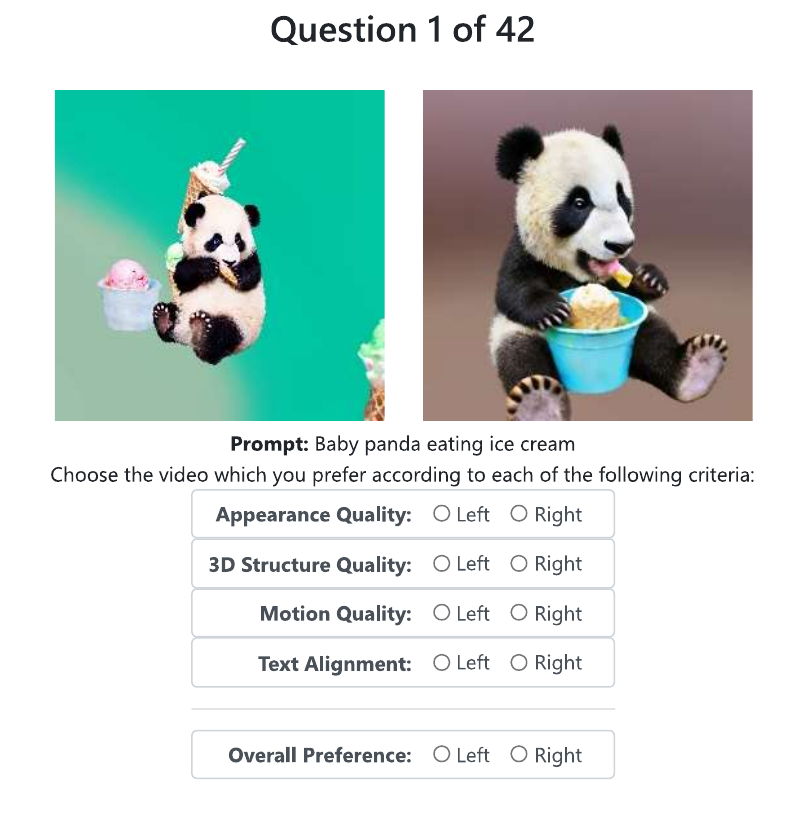}
\captionof{figure}{\textbf{Example survey question.} A survey question is shown above. The two videos are rendered using a baseline method (left) and 4D-fy (right). The left--right ordering of videos generated using each method is randomized throughout the survey.}
\label{fig:surveyquestion}
\end{figure}

The user study was carried out as a single survey consisting of 53 questions, each with 5 subquestions. Each question asked the evaluator to compare two videos: one showing a scene rendered with 4D-fy and another rendered with a separate method using the same text prompt as input. Evaluators filled out five subquestions that asked their preferred video based on appearance quality, 3D structure quality, motion quality, text alignment, and overall preference, as shown in Fig.~\ref{fig:surveyquestion}. Evaluators were given the following instructions for each metric.

\begin{itemize}
    \item \textbf{Appearance Quality:} Evaluate the clarity and visual appeal of the scene as it appears from any particular viewpoint (ignoring, e.g., inconsistencies in appearance across different viewpoints). Your assessment should focus on the appearance of the foreground object and ignore the background of the video.
    \item \textbf{3D Structure Quality:} Assess the detail and realism of the shape of the scene across the multiple viewpoints shown in the video. 
    \item \textbf{Motion Quality:} Assess the realism of motion, including the amount of motion and how naturally the movements in the video are portrayed. 
    \item \textbf{Text Alignment:} Determine how accurately each video reflects the content of the text prompt. Consider whether the key elements of the prompt are represented.
    \item \textbf{Overall Preference:} State your overall preference between the two videos. This is your subjective appraisal of which video, in your view, stands out as better based on appearance quality, 3D structure quality, motion quality, and text alignment, (i.e., overall quality).
\end{itemize}

Of the 53 questions, 28 were comparisons between 4D-fy and MAV3D. The other 25 questions were between 4D-fy and each of the five ablated methods described in Sec.~4.3 of the main paper, with five questions for each. For each comparison and metric, the results were tested against the null hypothesis that evaluators had no preference between either method; i.e., they would choose either with probability $0.5$. We aggregate over prompts and evaluators and use $\chi^2$ analysis to determine the corresponding $p$-value. We choose $p < 0.05$ as a significant deviation from the null hypothesis and find that---with the exception of comparing motion quality between 4D-fy and MAV3D---all $p$-values are well below $0.05$. %The $p$-value for text alignment and motion quality between 4D-fy and MAV3D are $0.0174$ and $0.3919$ respectively, while the rest are $< 0.0001$.
All statistically significant results indicated that users preferred videos rendered using 4D-fy over MAV3D.
\section{Qualitative Results}
\label{sec:supp_qualitative}

\begin{figure*}[t]
    \centering
    \includegraphics[width=6.9in]{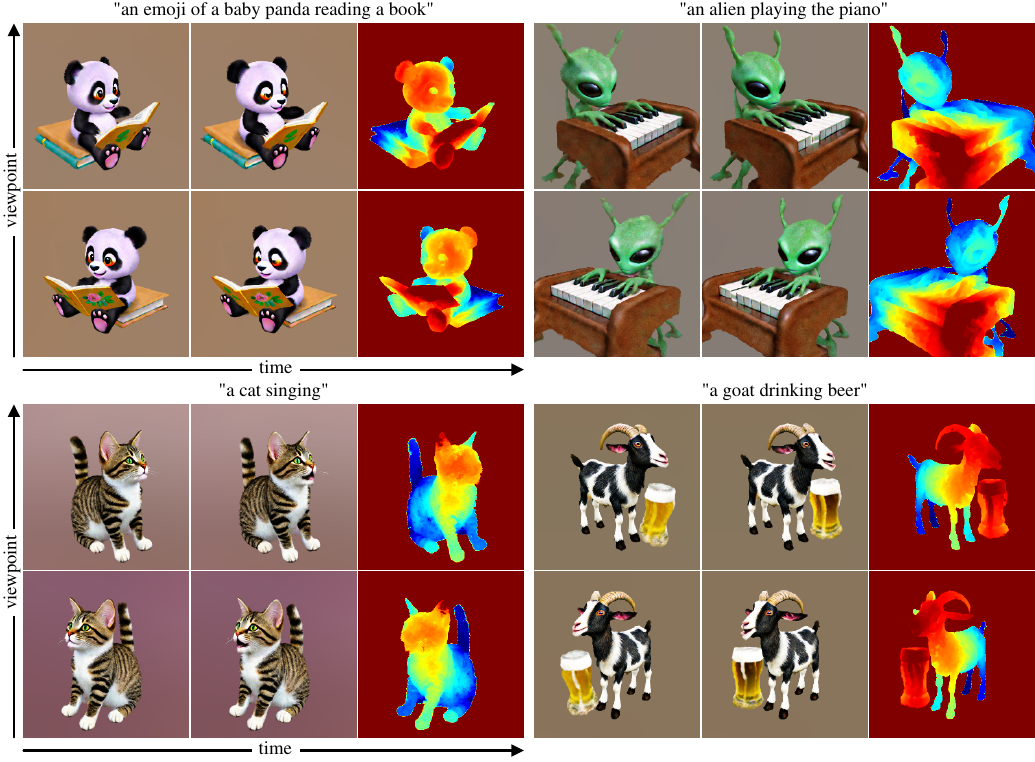}
    \captionof{figure}{\textbf{Text-to-4D Synthesis.} We present additional results using 4D-fy.}
\label{fig:supp_results}
\end{figure*}

In Fig. \ref{fig:supp_results}, we present additional results generated using our method. We highly encourage readers to view the videos included in our supplementary website to gain a better appreciation of our text-to-4D generation results.
\section{Ablations}
\label{sec:supp_ablations}

\begin{table}[!t]
    \vskip -0.05in
\caption{Further ablations: We indicate \% users who prefer 4D-fy over each method (Col.\ 1).}
    \vskip -0.2in
    \label{tab:rebuttal_results}
    \begin{center}
    \resizebox{\columnwidth}{!}{
    \begin{tabular}{lc|cccc|c}
        \toprule
        \textit{Method} & \textit{CLIP} & AQ & SQ & MQ & TA & Overall \\
        \midrule
        \textbf{4D-fy} & 35.0 & \multicolumn{4}{c|}{---} & ---\\
        
        w/ single-stage & 29.7 & 100\% & 100\% & 88\% & 97\% & 100\%\\
        w/ single hash grid & 25.5 & 100\% & 100\% & 99\% & 100\% & 100\% \\
        \bottomrule
    \end{tabular}}
    \end{center}
    \vskip -0.25in
\end{table}

\begin{figure*}[ht!]
  \centering
  \includegraphics[width=\textwidth]{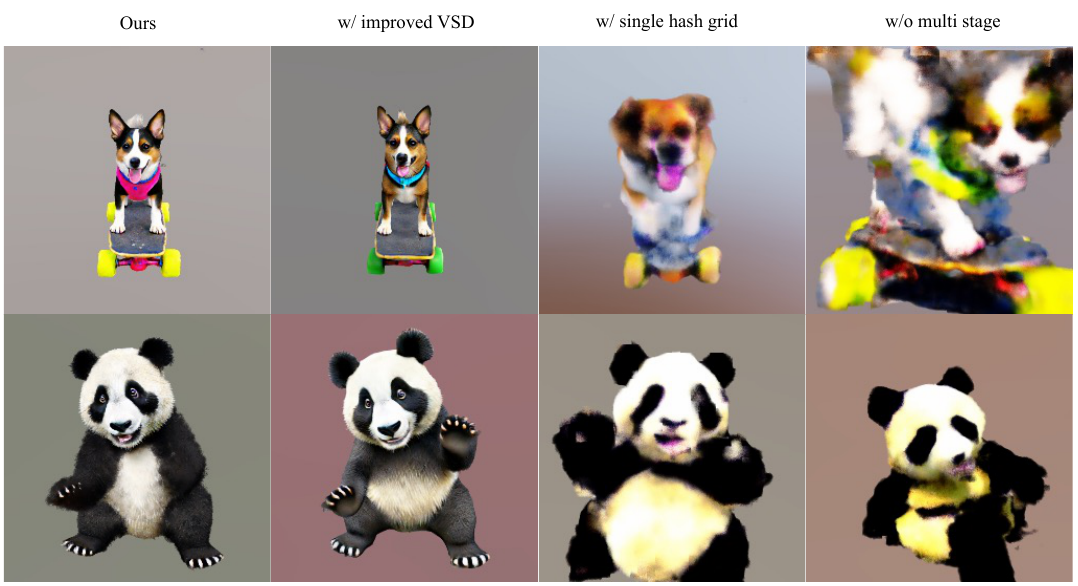}
  \vskip -0.1in
   \caption{Further ablation studies (zoom in for details). We also show that a slightly higher learning rate for VSD can improve the visual details.}
   \label{fig:ablations_extra}
   \vskip -0.20in
\end{figure*}

In Tab.~\ref{sec:supp_ablations} and Fig.~\ref{fig:ablations_extra} we show further ablation results. Training naively only the last stage using all three diffusion models leads to poor quality, as the model struggles to simultaneously learn 3D structure, appearance, and motion. Using a large single dynamic hash grid instead of decomposed smaller static and dynamic hash grids also leads to significantly lower quality results.
\section{Geometry}
\label{sec:supp_geometry}

\begin{figure*}[t!]
  \vskip -0.05in
  \centering
  \includegraphics[width=\textwidth]{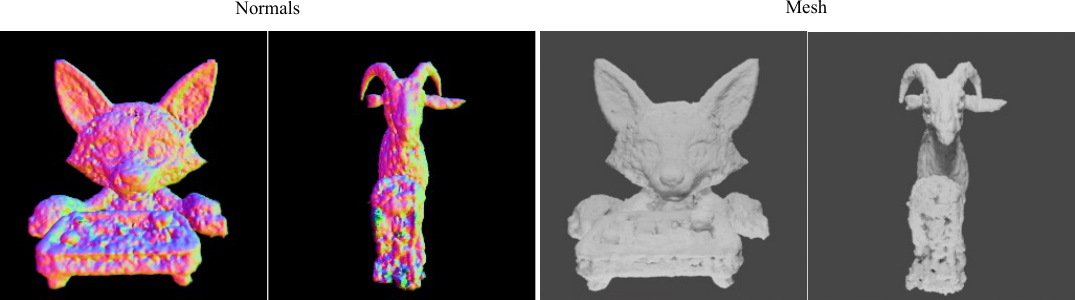}
  \vskip -0.1in
   \caption{Normals and meshes.}
   \label{fig:supp_geometry}
   \vskip -0.15in
\end{figure*}

In Fig.~\ref{fig:supp_geometry}, we show normals and meshes extracted with marching cubes. While generating high-quality geometry is not the main goal of this work, our method shows comparable quality to previous text-to-3D methods.
%Furthermore, we experimented with shading using normals and random lighting during training as introduced in DreamFusion~\cite{poole2022dreamfusion} and observe higher geometric quality at the cost of 15\% longer training. Generally, our model shows comparable geometric quality to text-to-3D methods.
\section{Limitations}
\label{sec:supp_limitations}

In addition to limitations outlined in the main paper, we briefly discuss the temporal flickering in renderings generated with 4D-fy. The main goal of our work is to generate high-quality dynamic 3D scenes---and especially to improve image quality compared to previous work where results can appear overly smooth or cartoon-like (e.g., Fig.~4). It may be possible to mitigate flickering artifacts by placing more emphasis on supervision with the video diffusion model; however, this may trade off image quality. Reducing temporal flickering without any penalty to image quality (e.g., avoiding blurry or overly smooth images) is an important direction for future work. 
\section{Ethics Statement}
\label{sec:supp_ethics}

In its current form, our method is not able to edit real people. However, it could be extended and misused for generating edited imagery of real people. We condemn the application of our method for creating realistic fake content intended to harm specific entities or propagate misinformation.

\end{document}